\documentclass[entropy,review,moreauthors,pdftex]{Definitions/mdpi} 

\setitemize{parsep=6pt,itemsep=0pt,leftmargin=*,labelsep=5.5mm,align=parleft}
\setenumerate{parsep=6pt,itemsep=0pt,leftmargin=*,labelsep=5.5mm,align=parleft}

\firstpage{1} 

\firstpage{1}  
 \pubvolume{xx} \issuenum{1} \articlenumber{1}
\pubyear{2018} \copyrightyear{2018}
\preto{\abstractkeywords}{\nolinenumbers}

\pdfoutput=1

\Title{High--Dimensional Brain in a High-Dimensional World: Blessing of Dimensionality}

\Author{{Alexander N. Gorban}{$^{1,2,}$}*\orcidA, 
Valery A. Makarov $^{2,3}$\orcidB   and Ivan Y. Tyukin $^{1,2}$\orcidC}

\AuthorNames{Alexander N. Gorban, Valery A. Makarov, Ivan Y. Tyukin}

\address{%
$^{1}$ \quad Department of Mathematics, University of Leicester, 
 Leicester LE1 7RH, UK; I.Tyukin@le.ac.uk\\
$^{2}$ \quad Laboratory of Advanced Methods for High-Dimensional Data Analysis, Lobachevsky University, {603022 Nizhny Novgorod}, Russia\\
$^{3}$ \quad Instituto de Matem\'{a}tica Interdisciplinar, Faculty of Mathematics, Universidad Complutense de Madrid, \mbox{Avda Complutense s/n}, 28040 Madrid, Spain; vmakarov@ucm.es}
\corres{Correspondence: {a.n.gorban@le.ac.uk}}

\abstract{High-dimensional data and high-dimensional representations of reality are inherent features of modern Artificial Intelligence systems and applications of machine learning. The well-known phenomenon of the ``curse of dimensionality'' states: many problems become exponentially difficult in high dimensions. Recently, the other side of the coin, the ``blessing of dimensionality'', has attracted much attention. It turns out that generic high-dimensional datasets exhibit fairly simple geometric properties. Thus, there is a fundamental tradeoff between complexity and simplicity in high dimensional spaces. Here we present a brief explanatory review of recent ideas,  results and hypotheses about the blessing of dimensionality and related simplifying effects relevant to machine learning and neuroscience.
}

\begin{document}


\section{Introduction}

 {During the last two decades, the curse of dimensionality in data analysis was complemented by the blessing of dimensionality: if a dataset is essentially high-dimensional then, surprisingly, some problems get easier and can be solved by simple and robust old methods. The curse and the blessing of dimensionality are closely related, like two sides of the same coin. The research landscape of these phenomena is gradually becoming more complex and rich. New theoretical achievements and applications provide a new context for old results. The single-cell revolution in neuroscience, phenomena of grandmother cells and sparse coding discovered in the human brain meet the new mathematical `blessing of dimensionality' ideas. In this mini-review, we aim to provide a short guide to new results on the blessing of dimensionality and to highlight the path from the curse of dimensionality to the blessing of dimensionality.  The selection of material and angle of view is based on our own experience. We are not trying to cover everything in the subject of review, but rather fill in the gaps in existing tutorials and~surveys.}

R. Bellman \cite{Bellman1957} in the preface to his book, discussed the computational difficulties of multidimensional optimization and summarized them under the heading ``curse of dimensionality''. He proposed to re-examine the situation, not as a mathematician, but as a ``practical man'' \cite{BellmanBAMS1954}, and concluded that the price of excessive dimensionality ``arises from a demand for too much information''. Dynamic programming was considered a method of dimensionality reduction in the optimization of a multi-stage decision process. Bellman returned to the problem of dimensionality reduction many times in different contexts~\cite{BellmanKalaba1961}. Now, dimensionality reduction is an essential element of the engineering (the ``practical man'') approach to  mathematical modeling \cite{GorbanEtAl2006}. Many model reduction methods were developed and successfully implemented in applications, from various versions of principal component analysis to approximation by manifolds, graphs, and complexes \cite{Joliffe2011,GorbanKegl2008,GorZin2010}, and low-rank tensor network decompositions~\cite{Cichocki2016,Cichocki2017}.

Various reasons and forms of the curse of dimensionality were classified and studied, from the obvious combinatorial explosion  (for example, for $n$ binary Boolean attributes, to check all the combinations of values we have to analyze $2^n$ cases) to more sophisticated distance concentration: in a high-dimensional space, the distances between randomly selected points tend to concentrate near their mean value, and the neighbor-based methods of data analysis become useless in their standard forms \cite{Beyer1999,Pestov2013}. Many ``good'' polynomial time algorithms become useless in high dimensions.

Surprisingly, however, and despite the expected challenges and difficulties, common-sense heuristics based on the simple and the most straightforward methods ``can yield results which are almost surely optimal'' for high-dimensional problems \cite{Kainen1997}. Following this observation, the term ``blessing of dimensionality'' was introduced \cite{Kainen1997,Brown1997}. It was clearly articulated as a basis of future data mining in the Donoho ``Millenium manifesto'' \cite{Donoho2000}. After that, the effects of the blessing of dimensionality were discovered in many applications, for example in face recognition \cite{Chen2013}, in analysis and separation of mixed data that lie on a union of multiple subspaces from their corrupted observations \cite{LiuG2016}, in multidimensional cluster analysis \cite{Murtagh2009}, in learning large Gaussian mixtures \cite{AndersonEtAl2014}, in correction of errors of multidimensonal machine learning systems \cite{GorbanTyuRom2016}, in evaluation of statistical parameters \cite{LiQ2018}, and in the development of generalized principal component analysis that provides low-rank estimates of the natural parameters by projecting the saturated model parameters \cite{Landgraf2019}.  

Ideas of the blessing of dimensionality became popular in signal processing, for example in compressed sensing \cite{Donoho2006,DonohoTanner2009} or in recovering a vector of signals from corrupted measurements \cite{candes2005}, and even in such specific problems as analysis and classification of EEG patterns for attention deficit hyperactivity disorder diagnosis \cite{Pereda2018}. 

There exist exponentially large sets of pairwise almost orthogonal vectors (`quasiorthogonal' bases,~\cite{Kurkova1993}) in Euclidean space.  It was noticed in the analysis of several $n$-dimensional random vectors drawn from the  standard Gaussian distribution with zero mean and identity covariance matrix, that all the rays from the origin to the data points have approximately equal length, are nearly orthogonal and   the distances between data points are all about $\sqrt{2}$ times larger \cite{Hall2005}. This observation holds even for exponentially large samples (of the size $\exp(a n)$ for some $a>0$, which depends on the degree of the approximate orthogonality) \cite{GorbTyuProSof2016}. Projection of a finite data set on random bases can reduce dimension with preservation of the ratios of distances (the Johnson–Lindenstrauss lemma  \cite{Dasgupta2003}).  

Such an intensive flux of works ensures us that we should not fear or avoid large dimensionality. We~just have to use it properly. Each application requires a specific balance between the extraction of important low-dimensional structures (`reduction') and the use of the remarkable properties of high-dimensional geometry that underlie statistical physics and other fundamental results  \cite{GorTyukPhil2018, Vershynin2018}. 

Both the curse and the blessing of dimensionality are the consequences of the measure concentration phenomena \cite{GianMilman2000,Ledoux2005,GorTyukPhil2018,Vershynin2018}. These phenomena were discovered in the development of the statistical backgrounds of thermodynamics. Maxwell, Boltzmann, Gibbs, and Einstein found that for many particles the distribution functions have surprising properties. For example, the Gibbs theorem of ensemble equivalence~\cite{Gibbs1902} states that a physically natural microcanonical ensemble (with fixed energy) is statistically equivalent (provides the same averages of physical quantities in the thermodynamic limit) to a maximum entropy canonical ensemble (the Boltzmann distribution).  Simple geometric examples of similar equivalence gives the `thin shell' concentration for balls: the volume of a high-dimensional ball is concentrated near its surface. Moreover, a high-dimensional sphere is concentrated near any equator (waist concentration; the general theory of such phenomena was elaborated by \mbox{M. Gromov \cite{Gromov2003}).} \mbox{P. L\'{e}vy \cite{Levy1951}} analysed these effects  and proved the first general concentration theorem. Modern measure concentration theory is a mature mathematical discipline with many deep results, comprehensive reviews  \cite{GianMilman2000}, books~\cite{Ledoux2005,Dubhashi2009}, advanced textbooks \cite{Vershynin2018}, and even elementary geometric introductions \cite{Bal1997}. Nevertheless, surprising counterintuitive results continue to appear and push new achievements in machine learning, Artificial Intelligence (AI), and neuroscience. 

This mini-review focuses on several novel results: stochastic separation theorems and evaluation of goodness of clustering in high dimensions, and on their applications to corrections of AI errors. Several possible applications to the dynamics of selective memory in the real brain and `simplicity revolution in neuroscience' are also briefly discussed.

\section{Stochastic Separation Theorems}
\unskip
\subsection{Blessing of Dimensionality Surprises and Correction of AI Mistakes}

D. Donoho and J. Tanner \cite{DonohoTanner2009} formulated several `blessing of dimensionality' surprises. In most cases, they considered $M$ points sampled independently from a standard  normal distribution  in dimension $n$. Intuitively, we expect that some of the  points will lie on the boundary of the convex hull of these points, and the others will be inside the interior of the hull. However, for large $n$ and $M$, this expectation is wrong. This is the main surprise. With a high probability $p>1-\varepsilon$ all $M$ random points are vertices of their convex hull. It is sufficient that $M<b\exp(a n)$ for some $a$ and $b$ that depend on $\varepsilon$ only \cite{GorbanGolubGrechTyu2018,GorbMakTyuk2019}. Moreover, with a high probability, each segment connecting a pair of vertices is also an edge of the convex hull, and  any simplex with $k$ vertices from the sample is a $k-1$-dimensional face of the convex hull for some range of values of $k$. For uniform distributions in a ball, similar results were proved earlier by I. {B{\'a}r{\'a}ny} and Z. {F{\"u}redi} \cite{convhull}. According to these results, each point of a random sample can be separated from all other points by a linear functional, even if the set is exponentially large.  

\begin{figure}[H]
\centering
\includegraphics[width=0.45\textwidth]{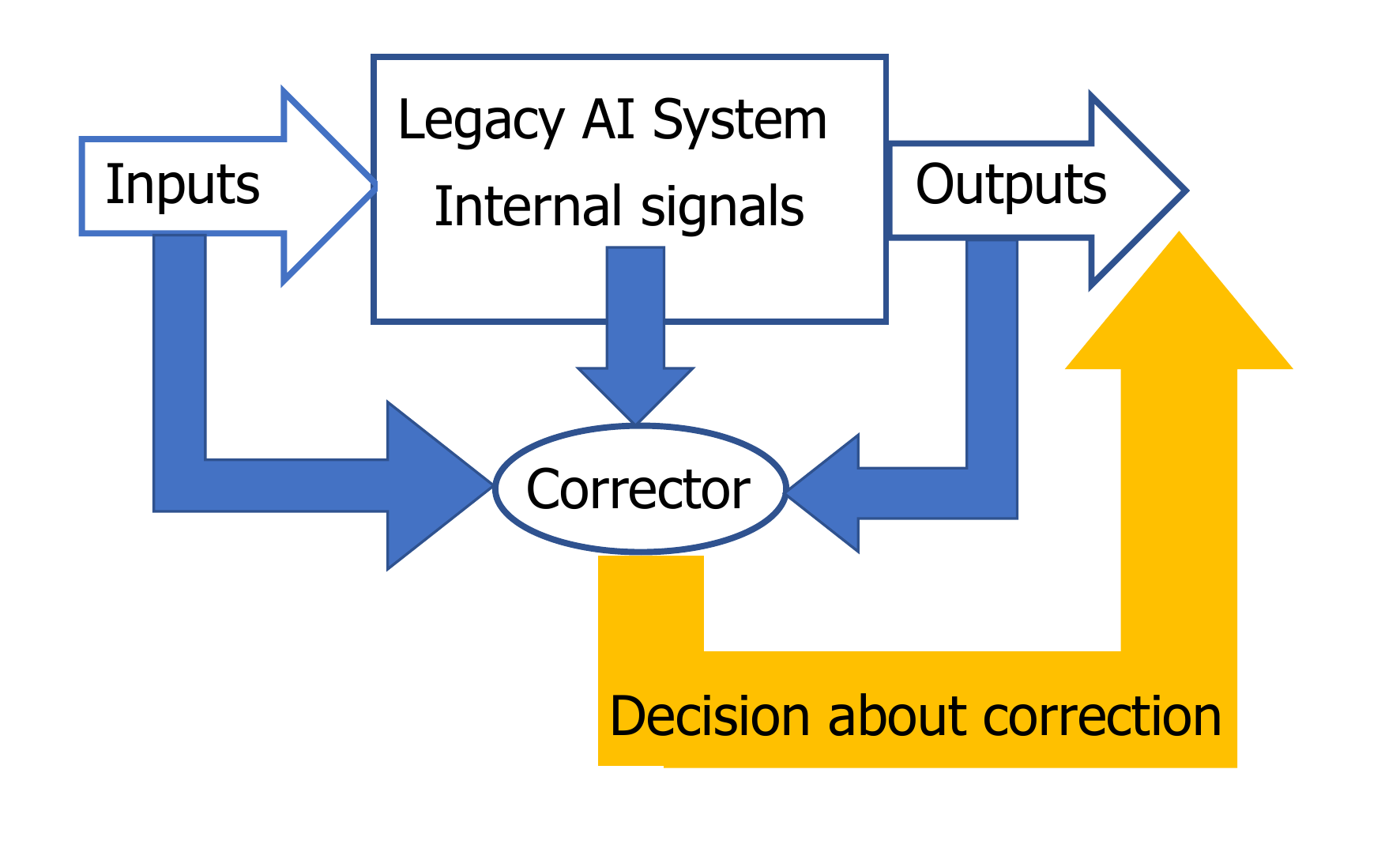}
\caption{A scheme of corrector. Corrector receives some input, internal, and output signals from the legacy artificial intelligence (AI) system and classifies the situation as `high risk' or `normal' one.  For a high-risk situation, it sends the corrected output to users following the correction rules. The high risk/normal situation classifier is prepared by supervised training on situations with diagnosed errors (universal construction). The online training algorithm could be very simple like Fisher's linear discriminants or their ensembles \cite{GorTyukPhil2018,GorbBurtRomTyu2019,GorbanGolubGrechTyu2018,GorbMakTyuk2019,TyukinGorMcEv}. Correction rules for high-risk situations are specific to a particular problem.}
\label{Fig:Corrector}
\end{figure}

Such a separability is important for the solution of a technological problem of fast, robust and non-damaging correction of AI mistakes \cite{GorTyukPhil2018,GorbanGolubGrechTyu2018,GorbMakTyuk2019}. AI systems make mistakes and will make mistakes in the future. If a mistake is detected, then it should be corrected. The complete re-training of the system requires too much resource and is rarely applicable to the  correction of a single mistake. We proposed to use additional simple machine learning systems,  correctors, for separation of  the situations with higher risk of mistake from the situations with normal functioning  \cite{GorbanTyuRom2016, GorbTyu2017} (Figure \ref{Fig:Corrector}). The decision rules  should be changed for situations with higher risk. Inputs for correctors are: the inputs of the original AI systems, the outputs of this system and (some) internal signals of this system \cite{GorbanGolubGrechTyu2018,GorbMakTyuk2019}. The construction of correctors for AI systems is crucial in the development of the future AI ecosystems.

Correctors should \cite{GorTyukPhil2018}: 
\begin{itemize}[leftmargin=*,labelsep=5.5mm]
\item be simple;  
\item not damage the existing  skills of the AI system; 
\item allow fast non-iterative learning;  
\item correct new mistakes without destroying the previous fixes. 
\end{itemize}

Of course, if an AI system made too many mistakes then their correctors could  conflict. In such a case, re-training is needed with the inclusion of new samples.

\subsection{Fisher Separablity}

Linear separation of data points from datasets \cite{convhull,DonohoTanner2009} is a good candidate for the development of AI correctors. Nevertheless, from the `practical man' point of view, one particular case, Fisher's discriminant~\cite{Fisher1936}, is much more preferable to the general case because it allows one-shot and explicit creation of the separating functional. 

Consider a finite data set $Y$ without any hypothesis about the probability distribution. Let {$(\,\cdot\, ,\, \cdot \,)$} 
be the standard inner product in $\mathbb{R}^n$.  {Let us define Fisher's separability following \cite{GorbanGolubGrechTyu2018}. }
\begin{Definition}
A point $\boldsymbol{ x}$ is Fisher-separable from a finite set $Y $ with a threshold $\alpha$ ($0\leq  \alpha<1$) if 
\begin{equation}\label{discriminant}
(\boldsymbol{x},\boldsymbol{y})\leq \alpha (\boldsymbol{x},\boldsymbol{x}), \;\;\;\mbox{  for all  }\boldsymbol{y}\in Y
\end{equation}
 \end{Definition}

This definition coincides with the textbook  definition of Fisher's discriminant if the data set $Y$ is whitened, which means  that the mean point is in the origin and the sample covariance matrix is the identity matrix. Whitening is often a simple by-product of principal component analysis (PCA) because, on the basis of principal components, whitening is just the normalization of coordinates to unit variance. Again, following the `practical' approach, we  stress that the precise PCA and whitening are not necessary but rather a priori  bounded condition number is needed: the ratio of the maximal and the minimal eigenvalues of the empirical covariance matrix after whitening should not exceed a given number $\kappa \geq 1$, independently of the dimension. 

A finite set is called Fisher-separable, if each point is Fisher-separable from the rest of the set  {(Definition 3, \cite{GorbanGolubGrechTyu2018}}).

\begin{Definition}A finite set $Y \subset {\mathbb R}^n$ is called {Fisher-separable} with threshold $\alpha \in (0,1)$ if inequality (\ref{discriminant})
holds for all $\boldsymbol{ x}, \boldsymbol{ y} \in F$ such that $\boldsymbol{ x}\neq  \boldsymbol{ y}$. The set $Y$ is called {Fisher-separable} if there exists some $\alpha $ ($0\leq  \alpha<1$)  such that $Y$ is Fisher-separable with threshold $\alpha$.  
\end{Definition}

Inequality (\ref{discriminant}) holds for vectors $\boldsymbol{x}$, $\boldsymbol{y}$ if and only if $\boldsymbol{x}$ does not belong to the ball (Figure~\ref{Fig:Excluded}):
\begin{equation}\label{excludedvolume}
\left\{\boldsymbol{z} \ \left| \ \left\|\boldsymbol{z}-\frac{\boldsymbol{y}}{2\alpha }\right\|< \frac{\|\boldsymbol{y}\|}{2\alpha} \right.  \right\}.
\end{equation}

\subsection{Stochastic Separation for Distributions with Bounded Support}

Let us analyse the separability of a random point from a finite set in the $n$-dimensional unit ball $\mathbb{B}_n$. Consider the distributions that can deviate from the equidistribution, and these deviations can grow with dimension $n$ but   not faster than the geometric progression with the common ratio $1/r>1$, and, hence, the maximal density $\rho_{\rm max}$ satisfies:
\begin{equation}\label{bounded}
\rho_{\rm max}<\frac{C}{r^n V_n(\mathbb{B}_n)},
 \end{equation}
where constant  $C$ does not depend on $n$.

\begin{figure}[H]
\centering
\includegraphics[width=0.6\textwidth]{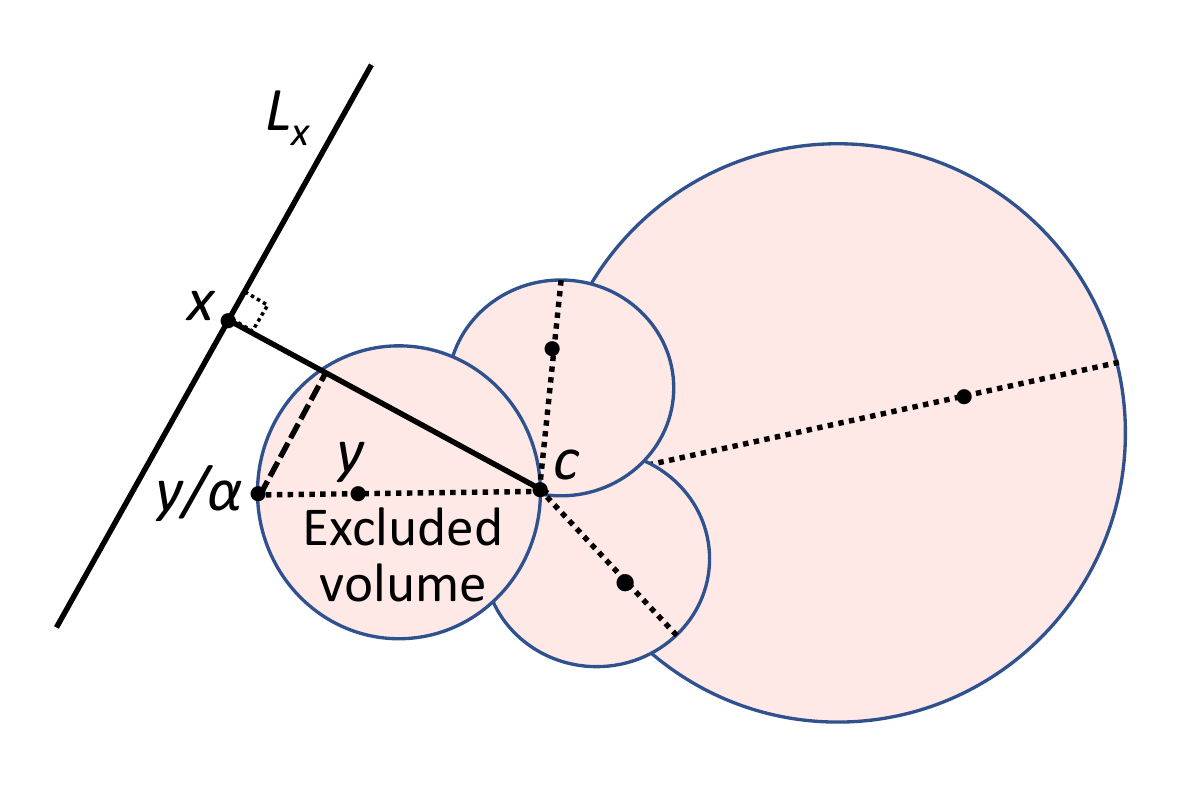}
\caption{{Fisher's separability of a point $\boldsymbol{x}$ from a set $Y$.} Diameters of the filled balls (excluded volume) are the segments $[\boldsymbol{c},\boldsymbol{y}/\alpha]$ ($y\in Y$). Point $\boldsymbol{x}$ should not belong to the excluded volume to be separable from $\boldsymbol{y}\in Y$ by the linear discriminant (\ref{discriminant}) with threshold $\alpha$. Here, $\boldsymbol{c}$ is the origin (centre), and $L_x=\{\boldsymbol{z}\, |\, (\boldsymbol{x},\boldsymbol{z})=(\boldsymbol{x},\boldsymbol{x})\}$ is the hyperplane.  A point $\boldsymbol{x}$ should not belong to the union of such balls (filled) for all $\boldsymbol{y} \in Y$ for separability from a set $Y$. The volume of this union,  $V_{\rm excl}$, does not exceed $ V_n(\mathbb{B}_n){|Y|}/{(2\alpha)^n}$.}
\label{Fig:Excluded}
\end{figure}

For such a distribution in the unit ball, the probability $\psi$ to find a random point   $\boldsymbol{x}$ in the excluded volume  $V_{\rm excl}$ (Figure \ref{Fig:Excluded}) tends to 0 as a geometric progression with the common ratio ${b}/({2r\alpha})$ when  $n\to \infty$.  

\begin{Theorem}\label{Theorem:ExclVol2}  {(Theorem 1, \cite{GorbanGolubGrechTyu2018})}
Let $Y \subset \mathbb{B}_n$, $|Y|<b^n$, and $2r\alpha>b>1$.
Assume that a probability distribution in the unit ball has a density with maximal value $\rho_{\rm max}$, which satisfies inequality (\ref{bounded}).
Then  the probability $p$ that a random point from this distribution is Fisher-separable from  $Y$ is $p=1-\psi$, where the probability of inseparability  $$\psi<C\left(\frac{b}{2r\alpha}\right)^n.$$
\end{Theorem}

Let us evaluate the probability that a random set $Y$ is Fisher-separable. Assume that each point of $Y$ is randomly selected from a distribution that satisfies  (\ref{bounded}). These distributions could be different for different $y\in Y$.

\begin{Theorem}\label{Theorem:ExclVol3}  {(Theorem 2, \cite{GorbanGolubGrechTyu2018})}
{Assume that a probability distribution} in the unit ball has a density with maximal value $\rho_{\rm max}$, which satisfies inequality (\ref{bounded}). Let $|Y|<b^n$ and $2r\alpha>b^2>1$.
Then  the probability $p$ that $Y$ is Fisher-separable  is $p=1-\psi$, where the probability of inseparability
 $$\psi<|Y|C\left(\frac{b}{2r\alpha}\right)^n<C\left(\frac{b^2}{2r\alpha}\right)^n.$$
\end{Theorem}

The difference in conditions from Theorem \ref{Theorem:ExclVol2} is that here $2r\alpha>b^2$ and in  Theorem \ref{Theorem:ExclVol2} $2r\alpha>b$.  Again,  $|Y|$ can grow exponentially with the dimension as the geometric progression with the common factor $b>0$, while $\psi \to 0$ faster than the geometric progression with the common factor ${b^2}/{2r\alpha}<1$.

For illustration, if $Y$ is an i.i.d. sample from the uniform distribution in the 100-dimensional ball and $|Y|=2.7\times 10^6$, then with probablity $p>0.99$ this set is Fisher-separable \cite{GorbTyu2017}.

\subsection{Generalisations}

V. K{\r{u}}rkov{\'a} \cite{KurkovaComm2019} emphasized that many attractive measure concentration results are formulated for i.i.d. samples from very simple distributions (Gaussian, uniform, etc.), whereas the reality of big data is very different: the data are not i.i.d. samples from simple distributions. The machine learning theory based on the i.i.d. assumption should be revised, indeed \cite{Symphony2019}. In the theorems above two main restrictions were employed: the probability of a set occupying relatively small volume could not be large (\ref{bounded}), and the support of the distribution is bounded. The requirement of identical distribution of different points is not needed. The independence of the data points can be relaxed \cite{GorbanGolubGrechTyu2018}. The boundedness  of the support of distribution can be transformed to the `not-too-heavy-tail' condition.  The condition `sets of relatively small volume should not have large probability' remains in most generalisations. It can be considered as `smeared absolute continuity' because  absolute continuity means that the sets of zero volume have zero probability. Theorems \ref{Theorem:ExclVol2} and \ref{Theorem:ExclVol3} have numerous generalisations \cite{GorbanGolubGrechTyu2018,GorbMakTyuk2019,Grechuk2019,KurkovaSang2019}. Let us briefly list some of~them:
\begin{itemize}[leftmargin=*,labelsep=5.5mm]
\item Product distributions in a unite cube where coordinates $X_i$ are independent random variables with the variances separated from zero, ${\rm var}(X_i)>\sigma_0^2>0$   {(Theorem 2, \cite{GorbTyu2017})}; significantly improved estimates are obtained by B. Grechuk \cite{Grechuk2019}.
\item Log-concave distributions (a distribution with density   $\rho(\boldsymbol{x})$  is log-concave if the set $D= \{\boldsymbol{x} | \rho(\boldsymbol{x})>0\}$ is convex and $g(\boldsymbol{x})=-\log \rho(\boldsymbol{x})$ is a convex  function on  $D$). In this case, the possibility of an exponential (non-Gaussian) tail brings a surprise: the upper size bound of the random set $|Y|$, sufficient for Fisher-separability in high dimensions with high probability, grows with dimension $n$ as $\sim \exp(a \sqrt{n})$, i.e. slower than exponential   {(Theorem 5, \cite{GorbanGolubGrechTyu2018})}.
\item Strongly log-concave distributions. A log concave distribution is strongly log-concave if   there exists a constant $c>0$ such that $$
\frac{g(\boldsymbol{x})+g(\boldsymbol{y})}{2} - g\left(\frac{\boldsymbol{x}+\boldsymbol{y}}{2}\right) \geq c \|\boldsymbol{x}-\boldsymbol{y}\|^2, \quad\quad \forall \boldsymbol{x},\boldsymbol{y} \in D.$$
In this case, we return to the exponential estimation of the maximal allowed size of $|Y|$  \mbox{{(Corollary 4, \cite{GorbanGolubGrechTyu2018})}}.
The comparison theorems \cite{GorbanGolubGrechTyu2018} allow us to combine different distributions, for example the distribution from Theorem \ref{Theorem:ExclVol3} in a ball with the log-concave or strongly log-concave tail outside the ball.
\item The kernel versions of the stochastic separation theorem were found, proved and applied to some real-life problems \cite{TyukinetalIJCNN2019}.
\item There are also various estimations beyond the standard i.i.d. hypothesis  \cite{GorbanGolubGrechTyu2018}  but the general theory is yet to be developed.
\end{itemize}

\subsection{Some Applications}

 {The correction methods were tested on various AI applications for videostream processing: detection of faces for security applications and detection of pedestrians \cite{GorbanGolubGrechTyu2018,SepTyukin2018,GorbBurtRomTyu2019}, translation of Sign Language into text for communication between deaf-mute people \cite{TyukinGreen2019}, knowledge transfer between AI systems~\cite{TyukinSofeikov2018}, medical image analysis, scanning and classifying archaeological artifacts \cite{AlisonArch2018}, etc., and even to some industrial systems with relatively high level of errors \cite{TyukinGorMcEv}.}

 {
Application of the corrector technology to image processing was patented together with industrial partners \cite{RomGorTyuPat}.
A typical test of correctors' performance is described below. For more detail of this test, we refer to  \cite{GorbBurtRomTyu2019}.  
A convolutional neural network (CNN) was trained to detect pedestrians in images. A set of 114,000 positive pedestrian and 375,000 negative non-pedestrian RGB images, re-sized to $128\times 128$, were collected and used as a training set. 
The testing set comprised of 10,000 positives and 10,000 negatives. The training and testing sets did not intersect. }

 {
We investigated in the computational experiments if it is possible to take one of cutting edge CNNs and train a one-neuron corrector to eliminate all the false positives produced. We also look at what effect, this corrector had on true positive numbers.}
  
 {For each positive and false positive  we extracted the second to last fully connected layer from CNN. These extracted feature vectors have dimension 4096. We applied PCA to reduce the dimension and analyzed how the effectiveness of the correctors depends on the number of principal components retained. This number varied in our experiments from 50 to 2000.
 The 25 false positives, taken from the testing set, were chosen at random to  model single mistakes of the legacy classifier. Several such samples were chosen.
For data projected on more than the first 87 principal components one neuron with weights selected by the Fisher linear discriminant formula corrected 25 errors without doing any damage to classification capabilities (original skills) of the legacy AI system on the training set. For 50 or less principal components this separation is not perfect.} 

 {Single false positives were corrected successfully without any increase of the true positive rates. We removed more than 10 false positives at no cost to true positive detections in the street video data (Nottingham) by the use of a single linear function. 
Further increasing the number of corrected false positives demonstrated that a single-neuron corrector could result in gradual deterioration of the true positive rates. }

\section{Clustering in High Dimensions}

Producing a special corrector for every single mistake seems to be a non-optimal approach, despite some successes. In practice, happily, often one corrector improves performance and prevents the system from some new mistakes because they are correlated. Moreover, mistakes can be grouped in clusters and we can create correctors for the clusters of situations rather than for single mistakes. Here we meet another measure concentration `blessing'. In high dimensions, clusters are good (well-separated) even in the situations when one can expect their strong intersection. For example, consider two clusters and the distance-based clustering. Let $r_1^2$ and $r^2_2$ be the mean squared Euclidean distances between the centroids of the clusters and their data points, and $\rho$ be the  distance between two centroids. The standard criteria of clusters' quality \cite{WunschClu2008} compare $\rho$ with $r_1+r_2$ and assume that for `good' clusters $r_1+r_2< \rho$. Assume the opposite, $r_1+r_2> \rho$ and evaluate the volume of the intersection of two balls with radii $r_1$, $r_2$. The intersection of the spheres (boundaries of the balls) is a $(n-1)$-dimensional sphere with the centre $c$  (Figure  \ref{Fig:BallIntersection}). Assume $\rho^2>|r_1^2-r_2^2|$, which means that $c$ is situated between the centers of the balls (otherwise, the biggest ball includes more than a half of the volume of the smallest one). The intersection of clusters belongs to a ball of radius $R$:

\begin{figure}[H]
\centering
\includegraphics[width=0.5\textwidth]{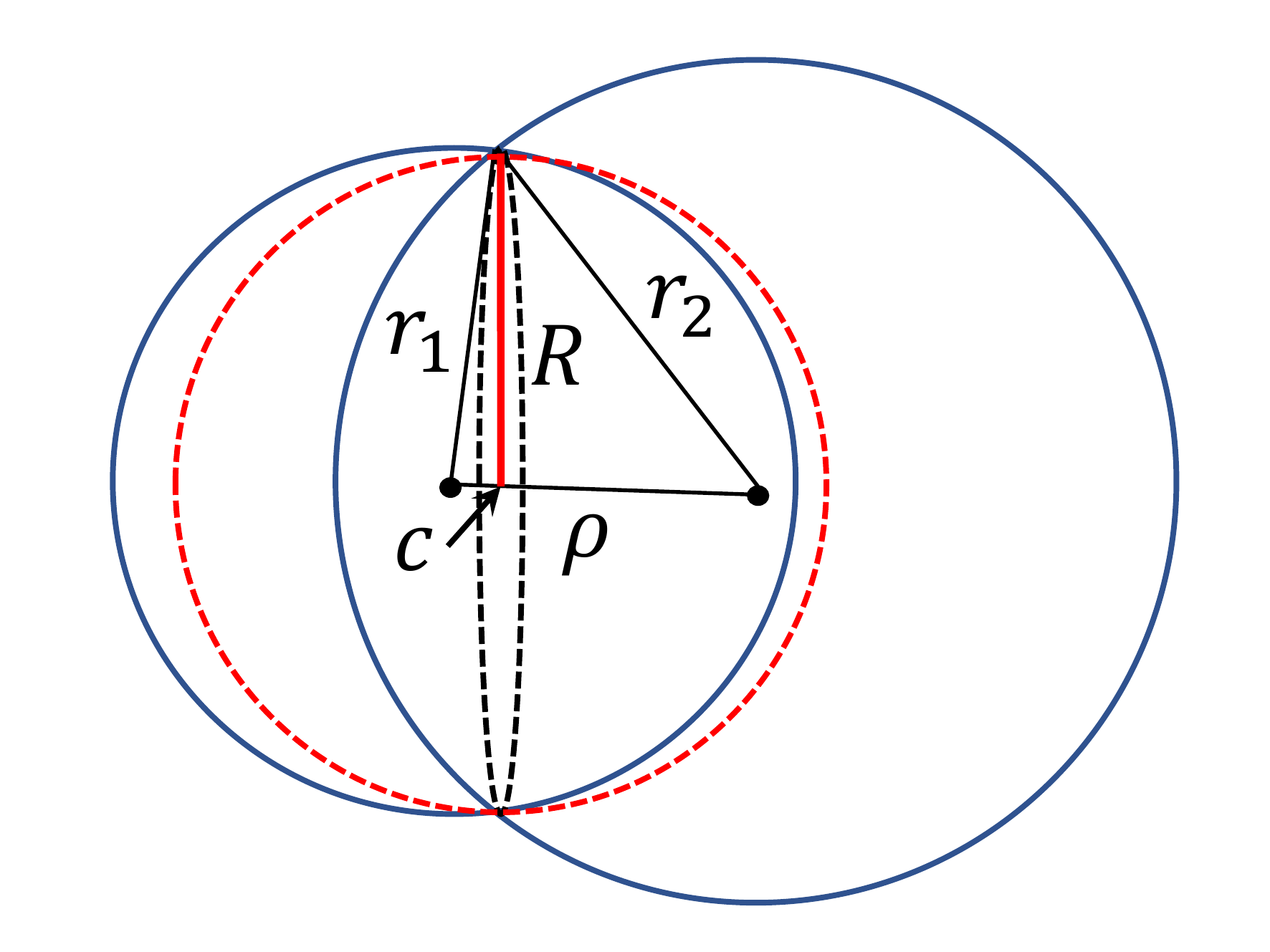} 
\caption{{Measure of clustering quality.} Intersection of two balls with the radii $r_1$, $r_2$ and the distance between centres $\rho<r_1+r_2$ is included in a ball with radius $R$ (\ref{BallIntersecR}) and centre $c$ (colored in red).}
\label{Fig:BallIntersection}
\end{figure}

\begin{equation}\label{BallIntersecR}
R^2=\frac{r_1^2+r_2^2}{2}-\frac{\rho^2}{4}-\frac{(r_1^2-r_2^2)^2}{4\rho^2}.
\end{equation}

$R<r_{1,2}$ and the fractions of the volume of the two initial balls in the intersection is less then $(R/r_{1,2})^n$. These fractions evaluate  the probability to confuse points between the clusters (for uniform distributions, for the Gaussian distributions the estimates are similar). We can measure the goodness of high-dimensional clusters by 
$$\gamma=\left(\frac{R}{r_1}\right)^n + \left(\frac{R}{r_2}\right)^n.$$

Note that $\gamma$ exponentially tends to zero with $n$ increase. Small $\gamma$ means `good' clustering. 

If  $\gamma \ll 1$ then the probability to find a data point in the intersection of the balls (the `area of confusion' between clusters) is negligible for uniform distributions in balls, isotropic Gaussian distributions and  always when small volume implies small probability. Therefore, the clustering of mistakes for correction of high-dimensional machine learning systems gives good results even if clusters are not very good in the standard measures,  and correction of clustered mistakes requires much fewer correctors for the same or even better accuracy \cite{TyukinGorMcEv}. 

 {We implemented the correctors with separation of clustered false-positive mistakes  from  the set of true positive  and tested them on the classical face detection task \cite{TyukinGorMcEv}. The legacy object detector was an OpenCV implementation of the Haar face detector. It has been applied to video footage capturing traffic and pedestrians  on the streets of Montreal.  The powerful MTCNN face detector was used to generate ground truth data. The total number of true positives was 21896, and the total number of false positives was 9372. The training set contained randomly chosen 50\%  of positives and false positives. PCA was used for dimensionality reduction with 200 principal components retained. Single-cluster corrector  allows one to filter 90\% of all errors at the cost of missing 5\% percent of true positives. In dimension 200, a cluster of errors is sufficiently well-separated from the true positives. A significant classification performance gain was observed with more clusters, up to 100. }

 {Further increase of dimension (the number of principal components retained) can even damage the performance because the number of features does not coincide with the dimension of the dataset, and the whitening with retained minor component can lead to ill-posed problems and loss of stability. For more detail, we refer to  
\cite{TyukinGorMcEv}.}

 {\section{What Does `High Dimensionality' Mean?}}

 {The dimensionality of data should not be naively confused with the number of features. Let us have $n$ objects with $p$ features. The usual {data matrix} in statistics is a 2D $n\times p$ array with $n$ rows and $p$ columns. The rows give values of features for an individual sample, and the columns give values of a  feature for different objects. In classical statistics, we assume that $n \gg p$ and even study asymptotic estimates for $n\to \infty$ and $p$ fixed. But, the modern `post-classical' world is different \cite{Donoho2000}: the situation with $n<p$ (and even  $n\ll p$) is not anomalous anymore.
Moreover, it can be considered in some sense as the generic case: we can measure a very large number of attributes for a relatively small number of individual cases.}

 {In such a situation the default preprocessing method could be recommended \cite{MoczkoMirkesGorPil2016}: transform the $n\times p$ data matrix with $n<p$  into the square $n\times n$ matrix of inner products (or correlation coefficients) between the individual data vectors. After that, apply PCA and all the standard machinery of machine learning. New data will be presented by projections on the old samples. (Detailed description of this preprocessing and the following steps is presented in \cite{MoczkoMirkesGorPil2016} with an applied case study for $n=64$ and $p\approx 5\times 10^5$.) Such a preprocessing reduces the apparent dimension of the data\_space to $p\leq n$.}

 {PCA gives us a tool for estimating the linear dimension of the dataset. Dimensionality reduction is achieved by using only the first few principal components.  
Several heuristics are used for evaluation of how many principal components should be retained:
\begin{itemize}[leftmargin=*,labelsep=5.5mm]
\item The classical Kaiser rule recommends to retain the principal components corresponding to the eigenvalues of the correlation matrix $\lambda \geq 1$ (or $\lambda \geq \alpha$ where $\alpha <1$ is a selected threshold; often $\alpha=0.5$ is selected). This is, perhaps, the most popular choice.
\item Control of the fraction of variance unexplained. This approach is also popular, but it can retain too many minor components that can be considered `noise'.
\item Conditional number control \cite{GorbanGolubGrechTyu2018} recommends to retain the principal components corresponding to $\lambda \geq \lambda_{\max}/\kappa$, where $\lambda_{\max}$ is the maximal eigenvalue of the correlation matrix and $\kappa$ is the upper border of the conditional number (the recommended values are $\kappa \leq 10$ \cite{Dormann2013}). This recommendation is very useful because it provides direct control of multicollinearity.
\end{itemize} }

 {After dimensionality reduction, we can perform whitening of data and apply the stochastic separation theorems. This requires a hypothesis about the distribution of data: sets of a relatively small volume should not have a high probability, and there should be no `heavy tails'. Unfortunately, this assumption is not always true in the practice of big data analysis. (We are grateful to G. Hinton and V. K{\r{u}}rkov{\'{a}} for this comment.)}

 {The separability properties can be affected by various violations of {i.i.d. }
 structure of data, inhomogeneity of data,  small clusters and fine-grained lumping, and other peculiarities \cite{AlberganteZinIJCNN2019}. Therefore, the notion of dimension should be revisited. We proposed to use the Fisher separability of data to estimate the dimension \cite{GorbanGolubGrechTyu2018}. For regular probability distributions, this estimate will give a standard geometric dimension, whereas, for complex (and often more realistic) cases, it will provide a more useful dimension characteristic. This approach was tested \cite{AlberganteZinIJCNN2019} for many bioinformatic datasets.}

 {For analysis of Fisher's separability and related estimation of dimensionality for  general distribution and empirical datasets, an auxiliary random variable is used \cite{GorbanGolubGrechTyu2018,AlberganteZinIJCNN2019}. This is the probability that  a randomly chosen point $\boldsymbol{x}$ is {not} Fisher-separable with threshold $\alpha$ from a given data point  $\boldsymbol{y}$ by the discriminant (\ref{discriminant}):}
\begin{equation}\label{excluded}
p=p_y(\alpha)=\int_{\left\|\boldsymbol{z}-\frac{\boldsymbol{y}}{2\alpha }\right\|\leq \frac{\|\boldsymbol{y}\|}{2\alpha}} \rho(\boldsymbol{z}) \,d\boldsymbol{z} ,
\end{equation}
where $ \rho(\boldsymbol{z}) \,d\boldsymbol{z}$ is the probability measure for  $\boldsymbol{x}$.
 
 {If $\boldsymbol{y}$ is selected at random (not compulsory with the same distribution as $\boldsymbol{x}$) then $p_y(\alpha)$ is a random variable. For a finite dataset $Y$ the probability $p_{Y}(\alpha)$ that the data point is {not}  Fisher-separable with threshold $\alpha$ from  $Y$ can be evaluated by the sum of $p_y(\alpha)$ for $\boldsymbol{y}\in Y$:
\begin{equation}\label{sump_y}
p_Y(\alpha) \leq \sum_{y\in Y} p_y(\alpha).
\end{equation}
 
 Comparison of the empirical distribution of  $p_y(\alpha)$ to the distribution evaluated for the high-dimensional sphere $\mathbb{S}^{n-1}\subset \mathbb{R}^{n}$  can be used as information about the `effective' dimension of data. The probability $p_y(\alpha)$ is the same for all $\boldsymbol{y} \in \mathbb{S}^{n-1}$ and exponentially decreases for large $n$. We assume that $\boldsymbol{y}$ is sampled randomly from for the rotationally invariant  distribution on the unit sphere  $\mathbb{S}^{n-1}\subset \mathbb{R}^{n}$. For large $n$ the asymptotic formula  holds  \cite{GorbanGolubGrechTyu2018,AlberganteZinIJCNN2019}:
\begin{equation}\label{p_y on sphere}
p_y(\alpha)\approx   \frac{(1-\alpha^2)^{(n-1)/2}}{\alpha \sqrt{2\pi (n-1)}}.
\end{equation}

Here $f(n)\approx g(n)$ means  that $f(n)/g(n)\to 1$ when $n\to \infty$ (the functions here are strictly positive). It was noticed that the asymptotically equivalent formula with the denominator $\alpha \sqrt{2\pi n}$ performs better in small dimensions \cite{AlberganteZinIJCNN2019}.}          

 {The introduced measure of dimension  performs  competitively with other state-of-the-art measures for simple  i.i.d. data situated on manifolds \cite{GorbanGolubGrechTyu2018,AlberganteZinIJCNN2019}. It was shown to perform better in the case of noisy samples and allows estimation of the intrinsic dimension in situations where the intrinsic manifold, regular distribution and i.i.d. assumptions are not valid \cite{AlberganteZinIJCNN2019}.
}

 {After this revision of the definition of data dimension, we can answer the question from the title of this section: What does `high dimensionality' mean? The answer is given by  the stochastic separation estimates for the uniform distribution in the unit sphere $\mathbb{S}^{n-1}\subset \mathbb{R}^{n}$. Let $y\in \mathbb{S}^{n-1}$. We use notation $A_m$ for the volume (surface) of $\mathbb{S}^m$. The points of $\mathbb{S}^{n-1}$, which are not Fisher-separable from $y$ with a given threshold $\alpha$, form a spherical cap with the base radius $r=\sqrt{1-\alpha^2}$ (Figure~\ref{CapSur}). The area of this cap is estimated from above by the lateral surface of the cone with the same base, which is tangent to the sphere at the base points  (see Figure~\ref{CapSur}). Therefore, the probability $\psi_{\alpha}$ that a point selected randomly from the rotationally invariant distribution on $\mathbb{S}^{n-1}$ is not Fisher-separable from $y$ is estimated from above as }
\begin{equation}
p_y(\alpha) < \frac{A_{n-2}}{A_{n-1}}\frac{(1-\alpha^2)^{(n-1)/2}}{\alpha (n-1)}. \label{psiAbove}
\end{equation}

The surface area of $\mathbb{S}^{n-1}$ is
\begin{equation}
A_{n-1}=\frac{2 \pi^{\frac{n}{2}}}{\Gamma\left(\frac{n}{2}\right)},
\end{equation}
where $\Gamma $ is Euler's gamma-function.

 {Rewrite the estimate (\ref{psiAbove}) as
\begin{equation}
p_y(\alpha) < \frac{\Gamma\left(\frac{n}{2}\right)}{\Gamma\left(\frac{n-1}{2}\right)}\frac{(1-\alpha^2)^{(n-1)/2}}{\alpha \sqrt{\pi}}. \label{psiAbove2}
\end{equation}

\begin{figure}[H]
\centering
\includegraphics[width=0.5\textwidth]{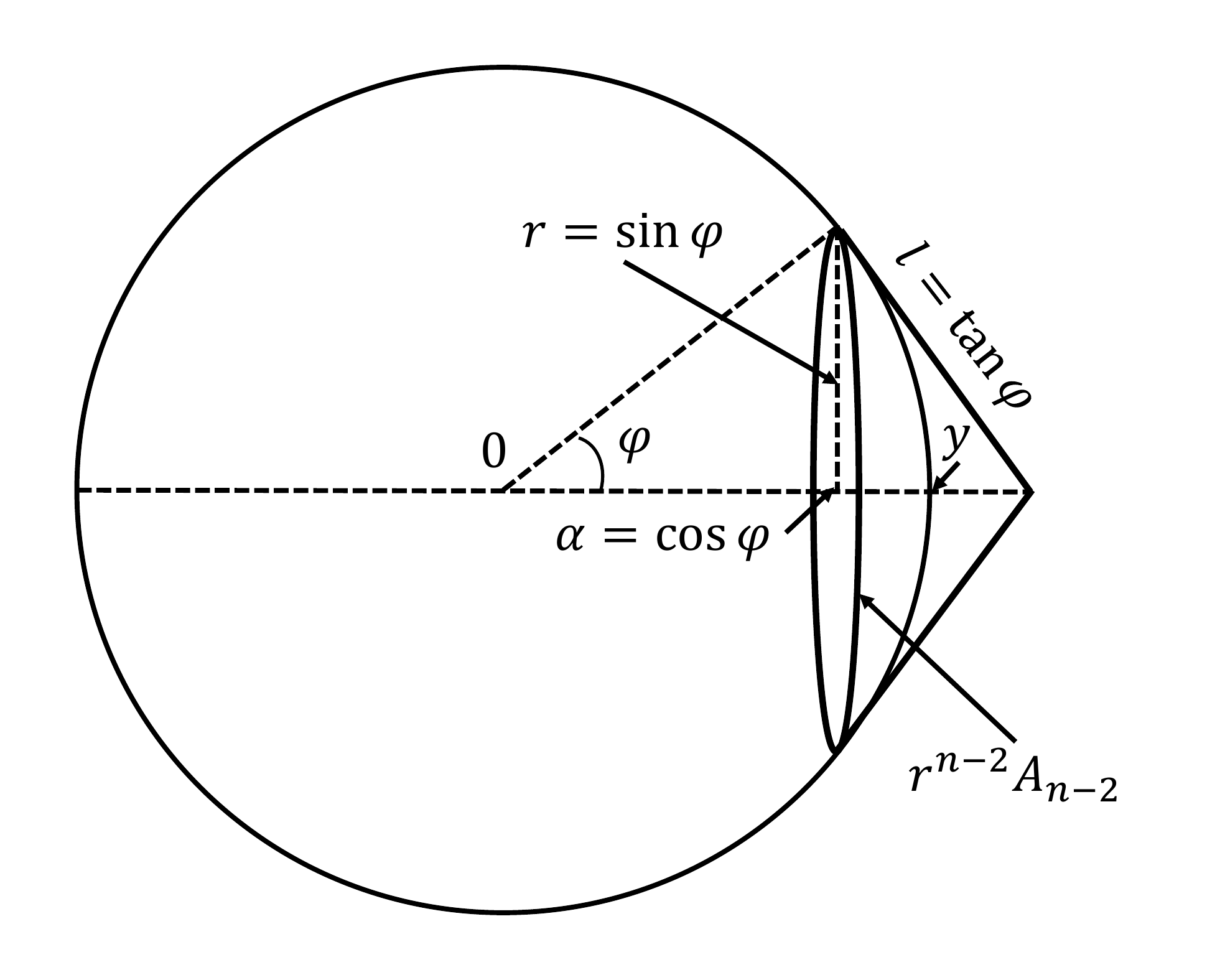} 
\caption{{Estimation of the area of the spherical cap.}  A point of $\mathbb{S}^{n-1}$ is Fisher-separable from $y\in \mathbb{S}^{n-1}$ with the threshold $\alpha=\cos \phi $ if and only if it  does not belong to the spherical cap   with the base radius $r=\sin \phi$ and  the base plane orthogonal to $y$. The surface of this spherical cap is less than the lateral surface of the cone that is tangent to the base. The $n-2$-dimensional surface of the base is $A_B=r^{n-2} A_{n-2}$. The lateral surface of the cone is $l A_B/(n-1)$. \label{CapSur}}
\end{figure}

Recall that $\Gamma(x)$ is a monotonically increasing logarithmically convex function for $x\geq 3/2$ \cite{Artin2015}. Therefore, for $n\geq 4$
$$\frac{\Gamma\left(\frac{n}{2}\right)}{\Gamma\left(\frac{n-1}{2}\right)}<
\frac{\Gamma\left(\frac{n+1}{2}\right)}{\Gamma\left(\frac{n}{2}\right)}.$$

Take into account that $\Gamma\left(\frac{n+1}{2}\right)=\frac{n-1}{2}\Gamma\left(\frac{n-1}{2}\right)$ (because $\Gamma(x+1)=x\Gamma(x)$). After elementary transforms it gives us
$$\frac{\Gamma^2\left(\frac{n}{2}\right)}{\Gamma^2\left(\frac{n-1}{2}\right)}<\frac{n-1}{2} \mbox{  and  }\frac{\Gamma\left(\frac{n}{2}\right)}{\Gamma\left(\frac{n-1}{2}\right)}< 
\frac{\sqrt{n-1}}{\sqrt{2}}.$$

 Finally, we got an elementary estimate for $p_y(\alpha)$ from above 
\begin{equation}
p_y(\alpha)< \frac{(1-\alpha^2)^{(n-1)/2}}{\alpha \sqrt{2\pi(n-1)}}. \label{psiAboveFin}
\end{equation}

Compared to (\ref{p_y on sphere}), this estimate from above is asymptotically exact. 
}
  
 {Estimate  from above a probability of a separability violations using (\ref{psiAboveFin}) and an elementary rule: for any family of events $U_1, U_2, \ldots , U_m$, 
\begin{equation}\label{probSum}
\mathbf{P}(U_1 \lor  U_2 \lor \ldots \lor U_m) \leq 
\mathbf{P}(U_1)+ \mathbf{P}(U_2)+\ldots + \mathbf{P}(U_m).
\end{equation}

According to (\ref{psiAboveFin}) and (\ref{probSum}), if $0<\psi<1$,  $Y$ is an i.i.d. sample from a rotationally invariant distribution on $\mathbb{S}^{n-1}$ and
\begin{equation}\label{EstimSphere1}
|Y|\frac{(1-\alpha^2)^{(n-1)/2}}{\alpha \sqrt{2\pi(n-1)}}<\psi,
\end{equation}
then all sample points with a probability greater than $1-\psi$ are Fischer-separable from a given point $ y $ with a threshold $\alpha$. Similarly, if 
\begin{equation}\label{EstimSphere2}
|Y|^2\frac{(1-\alpha^2)^{(n-1)/2}}{\alpha \sqrt{2\pi(n-1)}}<\psi,
\end{equation}
then with probability greater than $1-\psi$ each sample point is Fisher-separable from the rest of the sample with a threshold $\alpha$.}

 {Estimates (\ref{EstimSphere1}) and  (\ref{EstimSphere2}) provide sufficient conditions for separability. The Table \ref{Table:Sep} illustrates these estimates (the upper borders of $|Y|$ in these estimates are presented in the table with three significant figures). For an illustration of the separability properties, we estimated from above the sample size  for which the Fisher-separability is guaranteed with a probability 0.99 and a threshold value $ \alpha = 0.8 $ (Table  \ref{Table:Sep}). These sample sizes grow fast with dimension. From the Fisher-separability point of view, dimensions 30 or 50 are already large. The effects of high-dimensional stochastic separability emerge with increasing dimensionality much earlier than, for example, the appearance of exponentially large quasi-orthogonal bases \cite{GorbTyuProSof2016}.}

\begin{table}[H]
\centering
 {\caption{Stochastic separation on $n-1$-dimensional spheres. 
For a sample size less than $ M_{1,99} $, all points of an i.i.d. sample  with a probability greater than 0.99 are Fischer-separable from a given point $ y $ with a threshold $ \alpha = 0.8 $.
For a sample size less than $M_{2,99}$, with probability greater than 0.99  an i.i.d. sample is Fisher-separable with a threshold $\alpha=0.8$ (that is, each sample point is Fisher-separable from the rest of the sample with this threshold). \label{Table:Sep} }
\tablesize{\footnotesize} 
\begin{tabular}{ccccccccc}
\toprule
 	& \textbf{n = 10}	& \textbf{n = 20} & \textbf{n = 30} & \textbf{n = 40} & \textbf{n = 50} & \textbf{n = 60}& \textbf{n = 70}& \textbf{n = 80}\\
\midrule
$M_{1,99}$	& 5			& $1.43 \times 10^3$  & $2.94 \times 10^5$			& $5.91\times 10^7$ & $1.04\times 10^{10}$			& $1.89\times 10^{12}$ & $3.38\times 10^{14}$			& $5.98\times 10^{16}$    \\
$M_{2,99}$		& 2			& 37 & 542			& $7.49 \times 10^3$ & $1.02\times 10^5$			& $1.37\times 10^6 $  & $1.84 \times 10^7$			& $2.45 \times 10^8$  \\ 
\bottomrule
\end{tabular}}
\end{table}

\section{Discussion: The Heresy of Unheard-of Simplicity and Single Cell Revolution in Neuroscience}

V. Kreinovich \cite{Kreinovich2019} summarised the impression from the effective AI correctors based on Fisher's discriminant in high dimensions as ``The heresy of unheard-of simplicity''  using quotation of the famous Pasternak poetry. Such a simplicity appears also in brain functioning.  Despite our expectation that  complex intellectual phenomena is a result of a perfectly orchestrated collaboration between many different cells, there is a phenomenon of sparse coding, concept cells, or so-called `grandmother cells' which selectively react to the specific concepts like a grandmother or a well-known actress (`Jennifer Aniston cells') \cite{QuianQuirogaNature2005}. These experimental results continue the single neuron revolution  in sensory psychology \cite{Barlow1972}. 

The idea of grandmother or concept cells was proposed in the late 1960s. 
In 1972, Barlow  published a manifest about the single neuron
revolution in sensory psychology \cite{Barlow1972}. He suggested: ``our perceptions are caused by the activity of a rather small number of neurons selected from a very large population of predominantly silent cells.'' Barlow presented many experimental evidences of single-cell perception. In all these examples, neurons reacted selectively to the key patterns (called `trigger features'). This reaction was invariant to various changes
in conditions. 

The modern point of view on the single-cell revolution was briefly summarised recently  by \mbox{R. Quian Quiroga \cite{QuianQuiroga2019}}.   {He mentioned that the `grandmother cells' were invented by Lettvin ``to ridicule the idea that single neurons can encode specific concepts''. Later discoveries changed the situation and added more meaning and detail to these ideas. The idea of concept cells was evolved during decades. According to Quian Quiroga, these cells  are not involved in identifying a particular stimulus or concept. They are rather involved in creating and retrieving associations and can be seen as the ``building blocks of episodic memory''. Many recent discoveries used data received from intracranial
electrodes implanted in the medial temporal lobe (MTL; the hippocampus and surrounding cortex) for patients medications. The activity of dozens of neurons can be recorded while
patients perform different tasks. Neurons with high selectivity and invariance were found. 
In particular, one neuron fired to the presentation of seven different pictures of Jennifer Aniston and her spoken and written name, but not to 80 pictures of other persons.
Emergence of associations between images was also discovered.}

\begin{figure} [H]
\centering{\includegraphics[width = 0.75\textwidth]{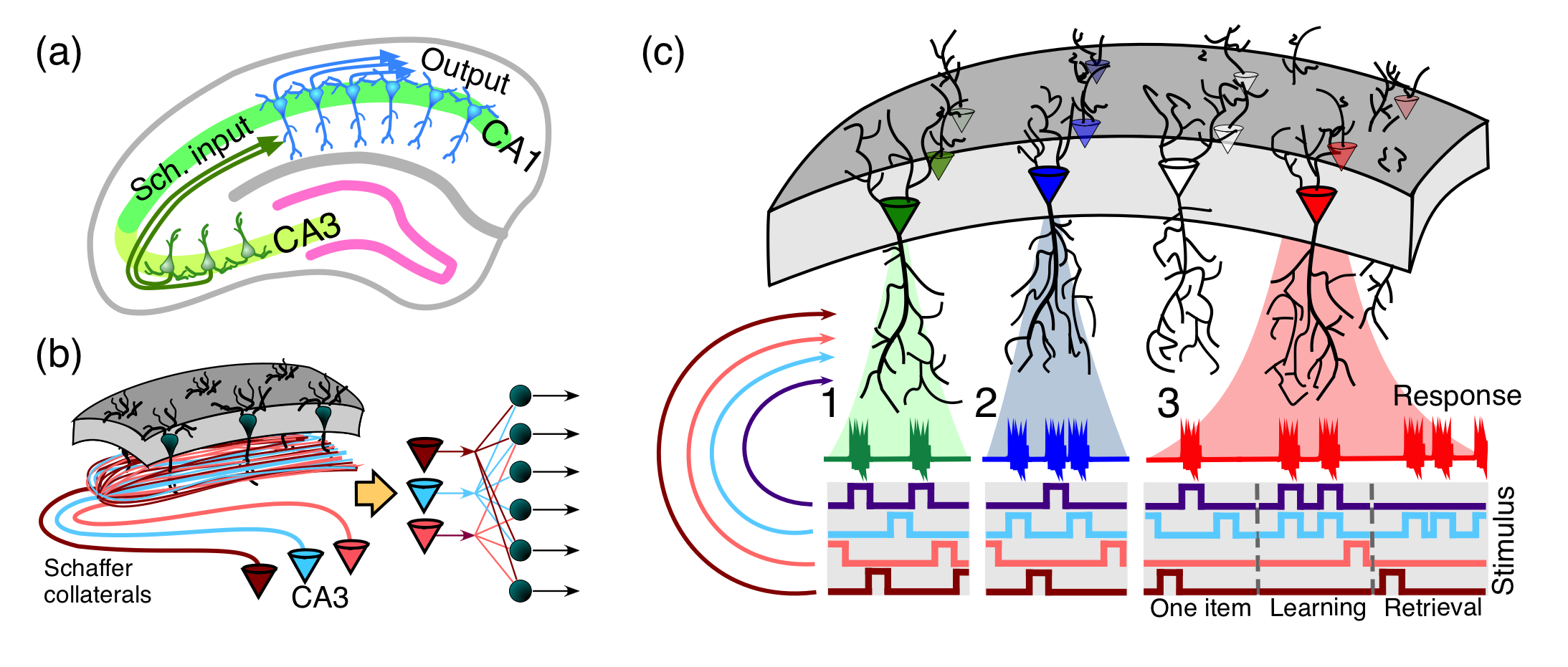}}
\caption{ {Organisation of encoding memories by single neurons in laminar structures: (\textbf{a}) laminar organization of the CA3 and CA1 areas in the hippocampus facilitates multiple parallel synaptic contacts between neurons in these areas by means of Schaffer collaterals; (\textbf{b}) axons from CA3 pyramidal neurons bifurcate and pass through the CA1 area in parallel (\textbf{left}) giving rise to the convergence-divergence of the information content (\textbf{right}). Multiple CA1 neurons receive multiple synaptic contacts from CA3 neurons; (\textbf{c}) schematic representation of three memory encoding schemes: (1) selectivity. A neuron ({shown} in green) receives inputs from multiple presynaptic cells that code different information items. It detects (responds to) only one stimulus (purple trace), whereas rejecting the others; (2) clustering. Similar to 1, but now a neuron ({shown} in blue) detects a group of stimuli (purple and blue traces) and ignores the others; (3) acquiring memories. A neuron (shown in red) learns dynamically a new memory item (blue trace) by associating it with a known one (purple trace). ((Figure 13, \cite{GorbMakTyuk2019}), published under CC BY-NC-ND 4.0 license.)}}
\label{FigNeurons}
\end{figure}

 {Some important memory functions are performed by stratified brain structures, such as the hippocampus.  The CA1 region of the hippocampus includes a monolayer of morphologically similar pyramidal cells oriented parallel to the main axis (Figure~\ref{FigNeurons}).  In humans, CA1 region of the hippocampus contains $ 1.4 \times 10^7$ of pyramidal neurons.  Excitatory inputs to these neurons come  from the CA3 regions (ipsi- and contra-lateral). Each CA3 pyramidal neuron sends an axon that bifurcates and leaves multiple collaterals in the CA1 (Figure~\ref{FigNeurons}b). 
This structural organization allows transmitting multidimensional information from the CA3 region to neurons in the CA1 region.  Thus, we have simultaneous convergence and divergence of the information content (Figure \ref{FigNeurons}b, right). A single pyramidal cell can receive  around 30,000 excitatory and 1700 inhibitory inputs (data for rats \cite{Megias2001}). Moreover, these numbers of synaptic contacts of cells vary greatly between neurons  \cite{Druckmann2014}. There are  nonuniform and clustered connectivity patterns. Such a variability is considered as a part of the mechanism enhancing neuronal feature selectivity~\cite{Druckmann2014}. However, anatomical connectivity is  not automatically transferred into functional connectivity and a realistic model  should decrease significantly (by several orders of magnitude) the number of functional connections (see, for example, \cite{Brivanlou2003}). Nevertheless, even several dozens of effective functional connections are sufficient for the application of stochastic separation theorems (see Table~\ref{Table:Sep}). }
  
For sufficiently high-dimensional sets of input signals a simple enough functional neuronal model with Hebbian learning (the generalized Oja rule \cite{TyuMak2018,GorbMakTyuk2019}) is capable of explaining the following phenomena: 
\begin{itemize}[leftmargin=*,labelsep=5.5mm]
\item the extreme selectivity of single neurons to the information content of high-dimensional data (Figure~\ref{FigNeurons}(c1)), 
\item simultaneous separation of several uncorrelated informational items from a large set of stimuli (Figure~\ref{FigNeurons}(c2)), 
\item dynamic learning of new items by associating them with already known ones (Figure~\ref{FigNeurons}(c3)). 
\end{itemize}

These results constitute a basis for the organization of complex memories in ensembles of single neurons. 

Re-training large ensembles of neurons is extremely time and resources consuming both in the brain and in machine learning.  It is, in fact, impossible to realize such a re-training in many real-life situations and applications. ``The existence of high discriminative units and a hierarchical organization for error correction are fundamental for effective information encoding, processing and execution, also relevant for fast learning and to optimize memory capacity'' \cite{VaronaComm2019}.

The multidimensional brain is the most puzzling example of the `heresy of unheard-of simplicity', but the same phenomenon has been observed in social sciences and in many other disciplines \cite{Kreinovich2019}.  

There is a fundamental difference and complementarity between analysis of essentially high-dimensional datasets, where simple linear methods are applicable, and reducible datasets for which non-linear methods are needed, both for reduction and analysis \cite{GorTyukPhil2018}. This alternative in neuroscience was described as high-dimensional `brainland' versus low-dimensional `flatland' \cite{BarrioComm2019}. The specific multidimensional effects of the `blessing of dimensionality'  can be considered as the deepest reason for the discovery of small groups of neurons that control important physiological phenomena. On the other hand, even low dimensional data live often in a higher-dimensional space and the dynamics of low-dimensional models should be naturally embedded into the high-dimensional `brainland'. Thus, a ``crucial problem nowadays is the `game' of moving from `brainland' to `flatland' and backward''  \cite{BarrioComm2019}.

C. van Leeuwen formulated a radically opposite point of view \cite{LeeuwenComm2019}: neither high-dimensional linear models nor low-dimensional non-linear models have serious relations to the brain.

The devil is in the detail. First of all, the preprocessing is always needed to extract the relevant features. The linear method of choice is PCA. Various versions of non-linear PCA can be also useful \cite{GorbanKegl2008}. After that, nobody has a guarantee that the dataset is either essentially high-dimensional or reducible. It can be a mixture of both alternatives, therefore both extraction of reducible lower-dimensional subset for nonlinear analysis and linear analysis of the high dimensional residuals could be needed together.

\authorcontributions{Conceptualization, ANG, VAM and IYT; Methodology, ANG, VAM and IYT; Writing--Original Draft Preparation, ANG; Writing--Editing, ANG, VAM and IYT; Visualization, ANG, VAM and IYT.}

\funding{The work was supported by the Ministry of Science and Higher Education of the Russian Federation  (Project No. 14.Y26.31.0022). Work of ANG and IYT was also supported by Innovate UK (Knowledge Transfer Partnership grants KTP009890; KTP010522) and University of Leicester. VAM acknowledges support from the Spanish Ministry of Economy, Industry, and Competitiveness (grant FIS2017-82900-P).  }

\conflictsofinterest{{The authors declare no conflict of interest.} } 
\reftitle{References}

\externalbibliography{yes}




\end{document}